\documentclass{article}

\PassOptionsToPackage{numbers, compress}{natbib}
\usepackage[final]{nips_2018}

\usepackage[utf8]{inputenc} 
\usepackage[T1]{fontenc}    
\usepackage{hyperref}       
\usepackage{url}            
\usepackage{booktabs}       
\usepackage{amsfonts}       
\usepackage{nicefrac}       
\usepackage{microtype}      
\usepackage{subfigure}
\usepackage{array,epsfig}
\usepackage{amsmath}      
\usepackage{amssymb}
\usepackage{amsxtra}
\usepackage{amsthm}
\usepackage{mathrsfs}
\usepackage{color}
\usepackage[normalem]{ulem}
\usepackage{caption}
\theoremstyle{definition}
\newtheorem*{rmk}{Remark}
\newtheorem{prop}{Proposition}

\title{Shared Representational Geometry Across Neural Networks}

\author{
Qihong Lu\\
Princeton University\\
\texttt{qlu@princeton.edu} \\
\And
Po-Hsuan Chen \\
Google Brain \\
\texttt{cameronchen@google.com} \\
\And
Jonathan W. Pillow \\
Princeton University \\
\texttt{pillow@princeton.edu} \\
\AND
Peter J. Ramadge \\
Princeton University \\
\texttt{ramadge@princeton.edu} \\
\And
Kenneth A. Norman \\
Princeton University \\
\texttt{knorman@princeton.edu} \\ 
\And 
Uri Hasson \\
Princeton University \\ 
\texttt{hasson@princeton.edu} \\ 
}

\begin{document}

\maketitle

\begin{abstract}
Different neural networks trained on the same dataset often learn similar input-output mappings with very different weights. Is there some correspondence between these neural network solutions? For linear networks, it has been shown that different instances of the same network architecture encode the same representational similarity matrix, and their neural activity patterns are connected by orthogonal transformations. However, it is unclear if this holds for non-linear networks. Using a shared response model, we show that different neural networks encode the same input examples as different orthogonal transformations of an underlying shared representation. We test this claim using both standard convolutional neural networks and residual networks on CIFAR10 and CIFAR100. 

\end{abstract}

\section{Introduction} 

Different people may share many cognitive functions (e.g. object recognition), but in general, the underlying neural implementation of these shared cognitive functions will be different across individuals. Similarly, when many instantiations of the same neural network architecture are trained on the same dataset, these networks tend to approximate the same mathematical function with very different weight configurations \cite{Dauphin2014-sq, Li2015-ur, Meng2018-vv}. 
Concretely, given the same input, two trained networks tend to produce the same output, but their hidden activity patterns will be different. In what sense are these networks similar? Broadly speaking, any mathematical function has many equivalent paramterizations. Understanding the connection of these paramterizations might help us understand the intrinsic property of that function. What is the connection across these neural networks trained on the same data? 


Prior research has shown that there are underlying similarities across the activity patterns from different networks trained on the same dataset \cite{Li2015-ur, Morcos2018-nf, Raghu2017-ng}. 
One hypothesis is that the activity patterns of these networks span highly similar feature spaces \cite{Li2015-ur}. Empirically, it has also been shown that different networks can be ``aligned'' by doing canonical correlation analysis on the singular components of their activity patterns \cite{Morcos2018-nf, Raghu2017-ng}. 
Interestingly, in the case of linear networks, prior theoretical research has shown that different instances of the same network architecture will learn the same representational similarity relation across the inputs \cite{saxe2014-ux, Saxe2018-bl}. 
And their activity patterns are connected by orthogonal transformations (assuming the training data is structured hierarchically, small norm weight initialization, and small learning rate) \cite{saxe2014-ux, Saxe2018-bl}. 
Though many conclusions derived from linear networks generalized to non-linear networks \cite{Advani2017-fo, saxe2014-ux, Saxe2018-bl}, it is unclear if this result holds in the non-linear setting. 

In this paper, we test if different neural networks trained on the same dataset learn to represent the training data as different orthogonal transformations of some underlying shared representation. 
To do so, we leverage ideas developed for analyzing group-level neuroimaging data. 
Recently, techniques have been developed for functionally aligning different subjects to a shared representational space directly based on brain responses \cite{Chen2015-mi, Haxby2011-uf}. 
Here, we propose to construct the shared representational space across neural networks with the shared response model (SRM) \cite{Chen2015-mi}, a method for functionally aligning neuroimaging data across subjects \cite{Anderson2016-xh, Guntupalli2016-so, Haxby2011-uf, Vodrahalli2018-kw}. 
SRM maps different subjects' data to a shared space through matrices with orthonormal columns. 
In our work, we use SRM to show that, in some cases, orthogonal matrices can be sufficient for constructing a shared representational space across activity patterns from different networks. 
Namely, different networks learn different rigid-body transformations of the same underlying representation. 
This result is consistent with the theoretical predictions made on deep linear networks \cite{saxe2014-ux, Saxe2018-bl}, as well as prior empirical works \citep{Li2015-ur, Morcos2018-nf, Raghu2017-ng}.

\begin{figure}
\centering
\begin{minipage}{.46\textwidth}
  \captionsetup{justification=centering}
  \centering
  \includegraphics[width=\linewidth]{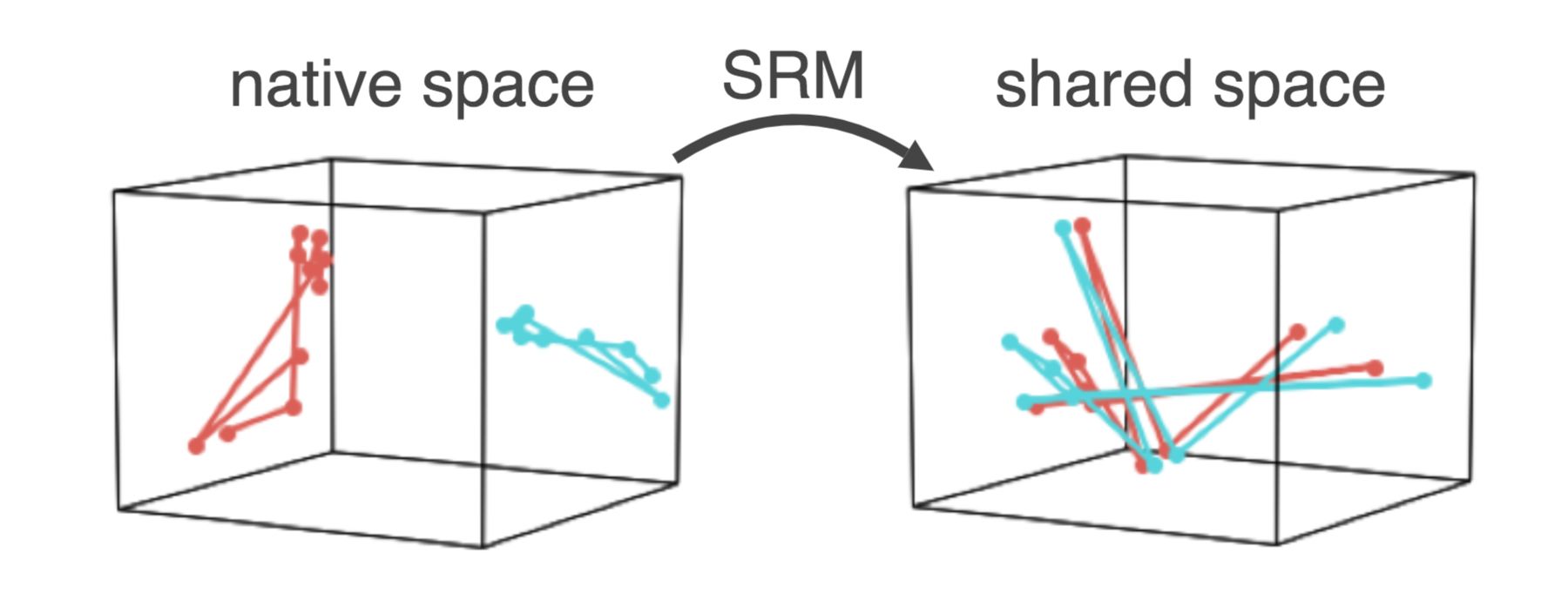}
  \caption*{A) SRM aligns activity patterns from different networks to a shared space}
\end{minipage}
\begin{minipage}{.52\textwidth}
  \captionsetup{justification=centering}
  \centering
  \includegraphics[width=\linewidth]{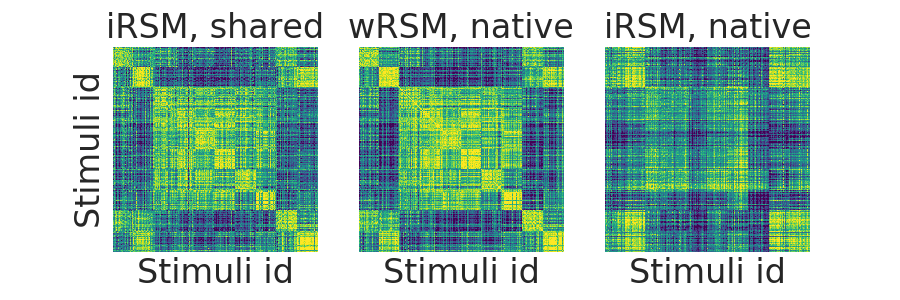}
  \caption*{B) In the shared space, inter-network RSM (iRSM) is similar to within-network RSM (wRSM)}
\end{minipage}
\caption{
{\bf{A)}} A low dimensional visualization of the hidden activity patterns of two networks. Each point is the average activity pattern of a class in CIFAR10. Before SRM, the hidden activity patterns on the same stimulus across the two networks seem distinct, but these patterns can be accurately aligned by orthogonal transformations. 
{\bf{B)}} Examples of shared space inter-network RSM (iRSM), native space within-network RSM (wRSM), and native space iRSM. In the shared space, iRSM is highly similar to the wRSM averaged across (ten) networks, suggesting the alignment is accurate. The native space iRSM is dissimilar from wRSM due to misalignment. 
}
\label{fig-demo}
\end{figure}

\section{Methods} 

Here we introduce the shared response model (SRM) and 
the concept of a 
representational similarity matrix (RSM). We 
use 
SRM to construct a shared representational space where hidden activity patterns across networks can be meaningfully compared. And we 
use
RSM to quantitatively evaluate the learned transformations. 

\textbf{Shared Response Model (SRM)}. 
SRM is formulated as in equation (\ref{eq:srm}). Given $N$ neural networks. Let $\mathbf{X}_i^l \in \mathbb{R}^{n \times m}$, be the set of activity patterns for $l$-th layer of network $i$, where $n$ is the number of units and $m$ is the number of examples. SRM seeks $\mathbf{S}_i^l \in \mathbb{R}^{k^l \times m}$, a basis set for the shared space, and $\mathbf{W}_i^l \in \mathbb{R}^{n \times k^l}$, the transformation matrices between the network-specific native space (the span of $\mathbf{X}_i^l$) and the shared space (Fig \ref{fig-demo}A shows a schematic illustration of this process). $\mathbf{W}_i^l$ are constrained to be matrices with orthonormal columns. Finally, $k^l$ is a hyperparameter that control the dimensionality of the shared space. When $k^l = n$, $\mathbf{W}_i^l$ is orthogonal, which represents a rigid-body transformation.
\begin{align}
    \label{eq:srm}
    \min_{\mathbf{W_i}^l, \mathbf{S}^l} 
    \sum_i || \mathbf{X_i}^l - \mathbf{W_i}^l \mathbf{S}^l||_F^2 
    \hspace{7pt} \text{s.t.}\hspace{7pt} 
    (\mathbf{W_i}^l)^T \mathbf{W_i}^l = \mathbf{I}_{k^l}
\end{align}

\textbf{Representational Similarity Matrix (RSM).} 
To assess the information encoded by hidden activity patterns, we use RSM \cite{Kriegeskorte2008-md, Kriegeskorte2013-xl}, a method for comparing neural representations across different systems (e.g. monkey vs. human). Let matrix $\mathbf{A} \in \mathbb{R}^{n \times m}$ to be the matrix of activity patterns for a neural network layer, where each column of $\mathbf{A}$ is an activity pattern evoked by an input. The within-network RSM of $\mathbf{A}$ is the correlation matrix of $\mathbf{A}$, 
i.e., $\text{RSM}_{ij}(\mathbf{A}) = \text{corr}(\mathbf{A}_i, \mathbf{A}_j)$. 
Without loss of generality, we assume $\mathbf{A}$ to be column-wise normalized, so $\text{RSM}(\mathbf{A}) = \mathbf{A}^T \mathbf{A}$. RSM is a $m \times m$ matrix that reflects all pairwise similarities of the hidden activity patterns evoked by different inputs. 
We define inter-network RSM as 
$\text{RSM}(\mathbf{A}, \mathbf{B}) = \mathbf{A}^T \mathbf{B}$. Figure \ref{fig-demo}B shows the RSMs from ten standard ConvNet trained on CIFAR10 for demonstration.

The averaged within-network RSM represents what's shared across networks. If two networks have identical activity patterns ($\mathbf{A} = \mathbf{B}$), their inter-network RSM will be identical to the averaged within-network RSM. 
However, if they are ``misaligned'' (e.g. off by an orthogonal transformation), their inter-network RSM will be different from the averaged within-network RSM. 
For example, consider two sets of patterns $\mathbf{A}$ and $\mathbf{B} = \mathbf{Q} \mathbf{A}$, where $\mathbf{Q}$ is orthogonal. 
Then 
$\text{RSM}(\mathbf{A}, \mathbf{B}) = \mathbf{A}^T \mathbf{B} = \mathbf{A}^T \mathbf{Q} \mathbf{A} \neq (\mathbf{A}^T \mathbf{A} + \mathbf{B}^T \mathbf{B}) / 2 = \text{RSM}(\mathbf{A})$. With this observation, we use the correlation between inter-network RSM and within-network RSM to assess the quality of SRM alignment.

\section{Results}

\textbf{The connection between SRM and representational similarity.} 
We start with establishing a theoretical connection between SRM and RSM -- if two sets of activity patterns $\mathbf{A}$, $\mathbf{B}$ have identical RSMs, $\mathbf{A}$, $\mathbf{B}$ can be represented as different orthogonal transformations of the same underlying shared representation.
Namely, there exist $\mathbf{W}_\mathbf{A} \in \mathbb{R}^{n \times k}$, $\mathbf{W}_\mathbf{B} \in \mathbb{R}^{n \times k}$ and $\mathbf{S}\in \mathbb{R}^{k \times m}$, such that $\mathbf{A} = \mathbf{W}_\mathbf{A} \mathbf{S} $ and $\mathbf{B} = \mathbf{W}_\mathbf{B} \mathbf{S} $, with $\mathbf{W}_\mathbf{A}^T \mathbf{W}_\mathbf{A} = \mathbf{I}_k$ and $\mathbf{W}_\mathbf{B}^T \mathbf{W}_\mathbf{B} = \mathbf{I}_k$. 
We prove this in the case of two networks, and the generalization to $N$ networks is straightforward.  

\prop{
For two sets of activity patterns $\mathbf{A}$ and $\mathbf{B}$, \text{RSM}($\mathbf{A}$) = \text{RSM}($\mathbf{B}$) if and only if $\mathbf{A}$ and $\mathbf{B}$ can be represented as different orthogonal transformations of the same shared representation $\mathbf{S}$.
\label{prop:srmrsa}
}

\textbf{Proof: }
For the forward direction, assume 
$\mathbf{A}^T\mathbf{A}  = \mathbf{B}^T \mathbf{B}$. 
Let 
$\mathbf{A} = \mathbf{U}_\mathbf{A} \mathbf{\Sigma}_\mathbf{A} \mathbf{V}_\mathbf{A}^T $
and 
$\mathbf{B} = \mathbf{U}_\mathbf{B} \mathbf{\Sigma}_\mathbf{B} \mathbf{V}_\mathbf{B}^T $ 
be compact SVDs. The assumption can be rewritten in terms of the SVDs: $\mathbf{V}_\mathbf{A} \mathbf{\Sigma}_\mathbf{A}^2 \mathbf{V}_\mathbf{A}^T = \mathbf{V}_\mathbf{B} \mathbf{\Sigma}_\mathbf{B}^2 \mathbf{V}_\mathbf{B}^T$. Under a generic setting, the eigenvalues will be distinct with probability one, so the two eigen-decompositions for corresponding covariance matrices are unique. Therefore, we have that
$\mathbf{\Sigma}_\mathbf{A} = \mathbf{\Sigma}_\mathbf{B}$ and $\mathbf{V}_\mathbf{A} = \mathbf{V}_B$. Let $\mathbf{\Sigma} := \mathbf{\Sigma}_\mathbf{A} = \mathbf{\Sigma}_\mathbf{B}$ and let $\mathbf{V} := \mathbf{V}_\mathbf{A} = \mathbf{V}_\mathbf{B}$. 
Now, we can rewrite $\mathbf{A}$ and $\mathbf{B}$ as $\mathbf{A} = \mathbf{U}_\mathbf{A} \mathbf{\Sigma} \mathbf{V}^T $ and $\mathbf{B} = \mathbf{U}_\mathbf{B} \mathbf{\Sigma} \mathbf{V}^T$. Finally, let $\mathbf{W}_\mathbf{A} = \mathbf{U}_\mathbf{A}$, $\mathbf{W}_\mathbf{B} = \mathbf{U}_\mathbf{B}$, and $\mathbf{S} = \mathbf{\Sigma} \mathbf{V}^T$. By construction, this is a SRM solution that perfectly aligns $\mathbf{A}$ and $\mathbf{B}$. 

For the converse, assuming there is a SRM solution that achieves a perfect alignment for $\mathbf{A}$ and $\mathbf{B}$. Namely, $\mathbf{A} = \mathbf{W}_\mathbf{A} \mathbf{S}$ and $\mathbf{B} = \mathbf{W}_\mathbf{B} \mathbf{S}$, with $\mathbf{W}_\mathbf{A}^T \mathbf{W}_\mathbf{A} = \mathbf{I}_k$ and $\mathbf{W}_\mathbf{B}^T \mathbf{W}_\mathbf{B} = \mathbf{I}_k$ for some $\mathbf{S}, \mathbf{W}_\mathbf{A}, \mathbf{W}_\mathbf{B}$. Then, 
\begin{align}
\label{eq:prop1_rev}
\mathbf{A}^T \mathbf{A} 
= (\mathbf{W}_\mathbf{A}\mathbf{S})^T \mathbf{W}_\mathbf{A}\mathbf{S} 
= \mathbf{S}^T \mathbf{W}_\mathbf{A}^T \mathbf{W}_\mathbf{A} \mathbf{S} 
= \mathbf{S}^T \mathbf{S} 
= ...
= \mathbf{B}^T \mathbf{B}  
\end{align}
\qed

\rmk{Equation \ref{eq:prop1_rev} made clear that if SRM can align activity patterns across two networks perfectly, then $\mathbf{S}$ will capture the RSM, or the covariance of the activity patterns, as $\mathbf{S}^T \mathbf{S} = \mathbf{A}^T \mathbf{A} = \mathbf{B}^T \mathbf{B}$.
}


\begin{figure}[!t]
    \begin{minipage}{.49\textwidth}
    \centering
    \includegraphics[width=\linewidth]{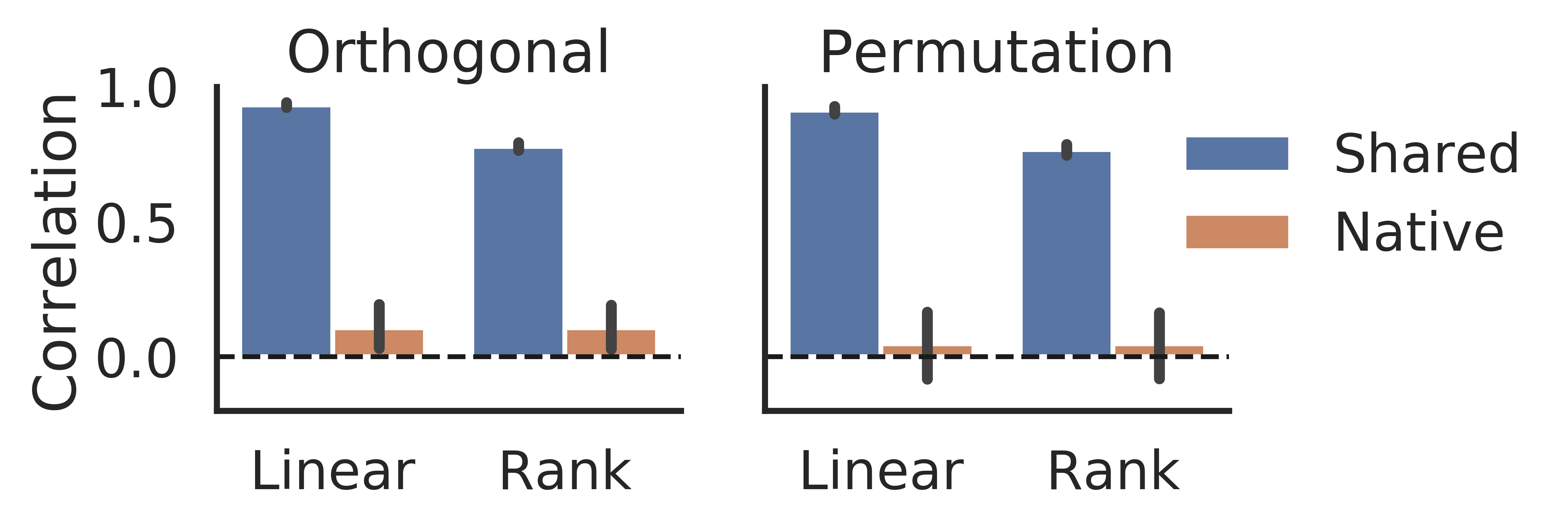}
    \caption*{A) Reconstruct the shared representation}
    \end{minipage}
    \begin{minipage}{.49\textwidth}
    \centering
    \includegraphics[width=\linewidth]{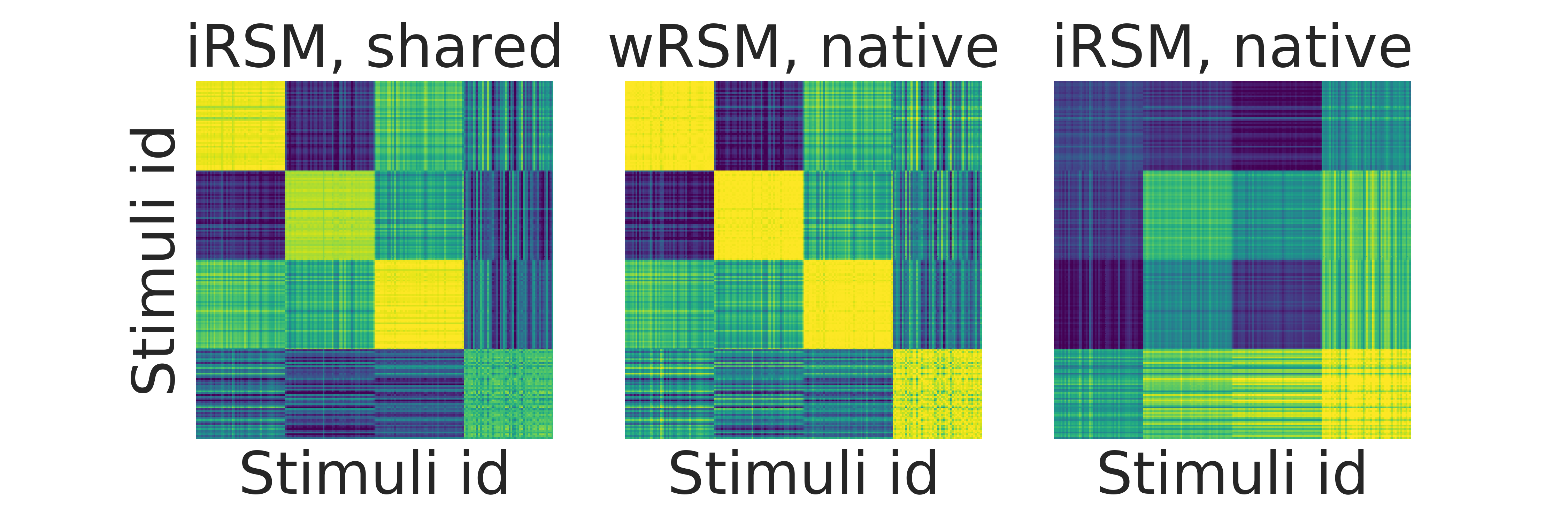}
    \caption*{B) Example RSMs from one simulation}
    \end{minipage}
    \caption{
    SRM can reconstruct a shared space that captures the underlying shared RSM. 
    \textbf{A)} Synthetic activity patterns for ten networks are generated by multiplying the activity patterns of a trained network by ten random orthogonal matrices. The bar plot shows the average correlation between the averaged within-network RSM (wRSM) and the inter-network RSM (iRSM) in the shared space (blue) and the native space (brown). The results based on linear vs. rank correlation are highly similar. 
    The error bars show 95\% bootstrapped confidence interval based on 50 simulation runs. 
    \textbf{B)} Visualization of the average wRSM in the native space and average iRSM in the shared space and native spaces. In the shared space, iRSM is highly similar to the wRSM. 
    }
    \label{fig-sim}
\end{figure}

\textbf{Simulation: SRM can identify orthogonal transformations.}
Proposition \ref{prop:srmrsa} shows that if two sets of activity patterns have the same RSM, then their connection is in the search space of SRM. 
However, there is no guarantee that SRM can identify such connection, as SRM objective is non-convex. In this simulation, we test whether SRM can align different sets of activity patterns, artificially created so that they that are connected by orthogonal transformations. 

Specifically, we trained a neural network on a toy classification task (not linearly separable). 
We then recorded its hidden activity matrix, $\mathbf{H} \in \mathbb{R}^{n \times m}$, where $n$ is the number of hidden units and $m$ is the number of examples, on unseen test points. To generate the synthetic activity patterns, 
we multiply $\mathbf{H}$ by ten random orthogonal (or permutation) matrices, $\mathbf{Q}_i, i = 1, ..., 10$. 
We then divided $\mathbf{Q}_i \mathbf{H}$ as a SRM-alignment set and a test set. We trained SRM on the SRM-alignment set and used the learned transformations to transform the test set activity patterns. All later analyses were conducted on the test set activity patterns. 

The goal is to evaluate whether SRM can construct a shared space such that the set of activity patterns, $\mathbf{Q}_i \mathbf{H}, \forall i$ are well aligned. The quality of alignment is measured by the correlation between the average inter-network RSM and the average within-network RSM. 
Figure \ref{fig-sim}A shows that the average inter-network RSM in the shared space is highly similar to the average within-network RSM in the native space. This demonstrates that SRM can construct a shared space that aligns activity patterns across different networks, if their activity patterns are connected by orthogonal transformations. Figure \ref{fig-sim}B shows several RSMs from one simulation for demonstration. 

\begin{figure}[t!] 

  \begin{minipage}[b]{0.24\linewidth}
    \centering
    \includegraphics[width=\linewidth]{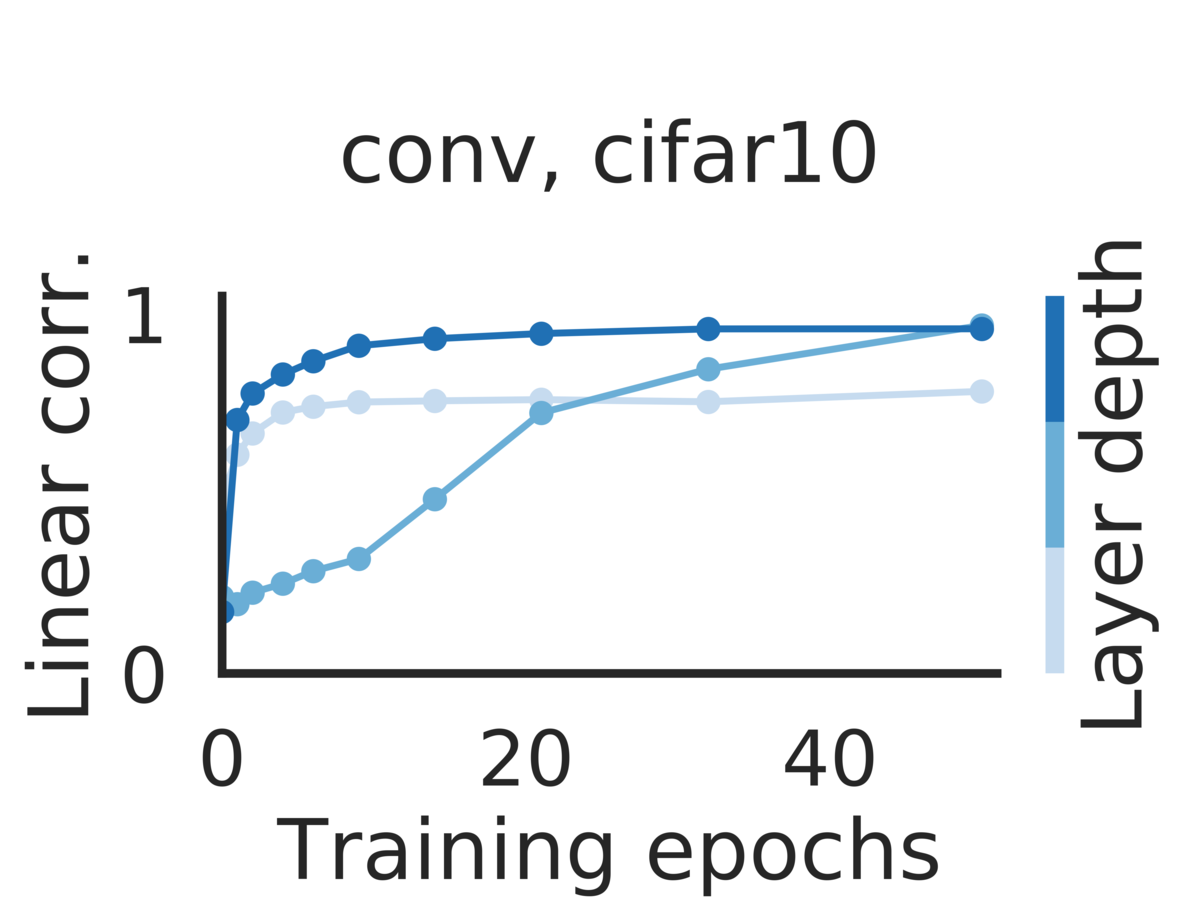}
  \end{minipage}
  \begin{minipage}[b]{0.24\linewidth}
    \centering
    \includegraphics[width=\linewidth]{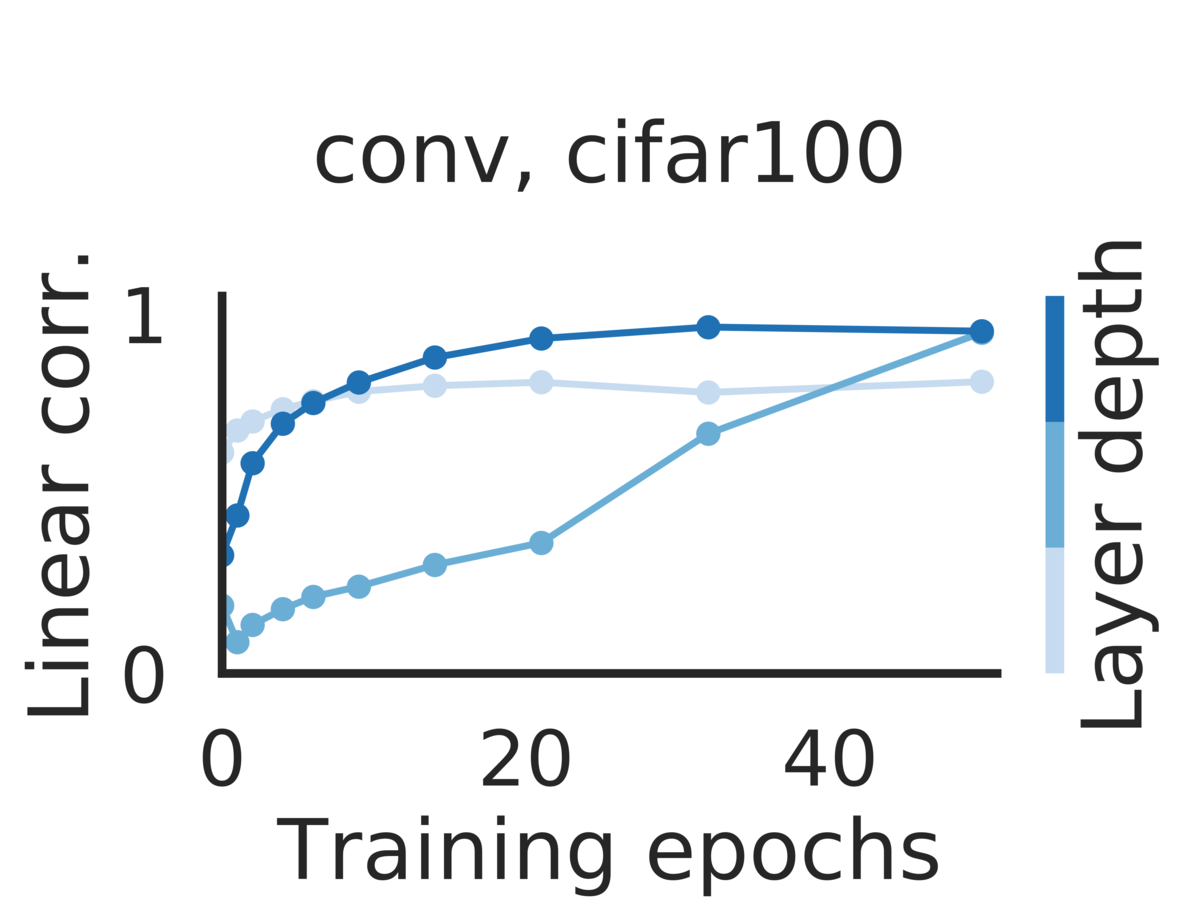}
  \end{minipage} 
  \begin{minipage}[b]{0.24\linewidth}
    \centering
    \includegraphics[width=\linewidth]{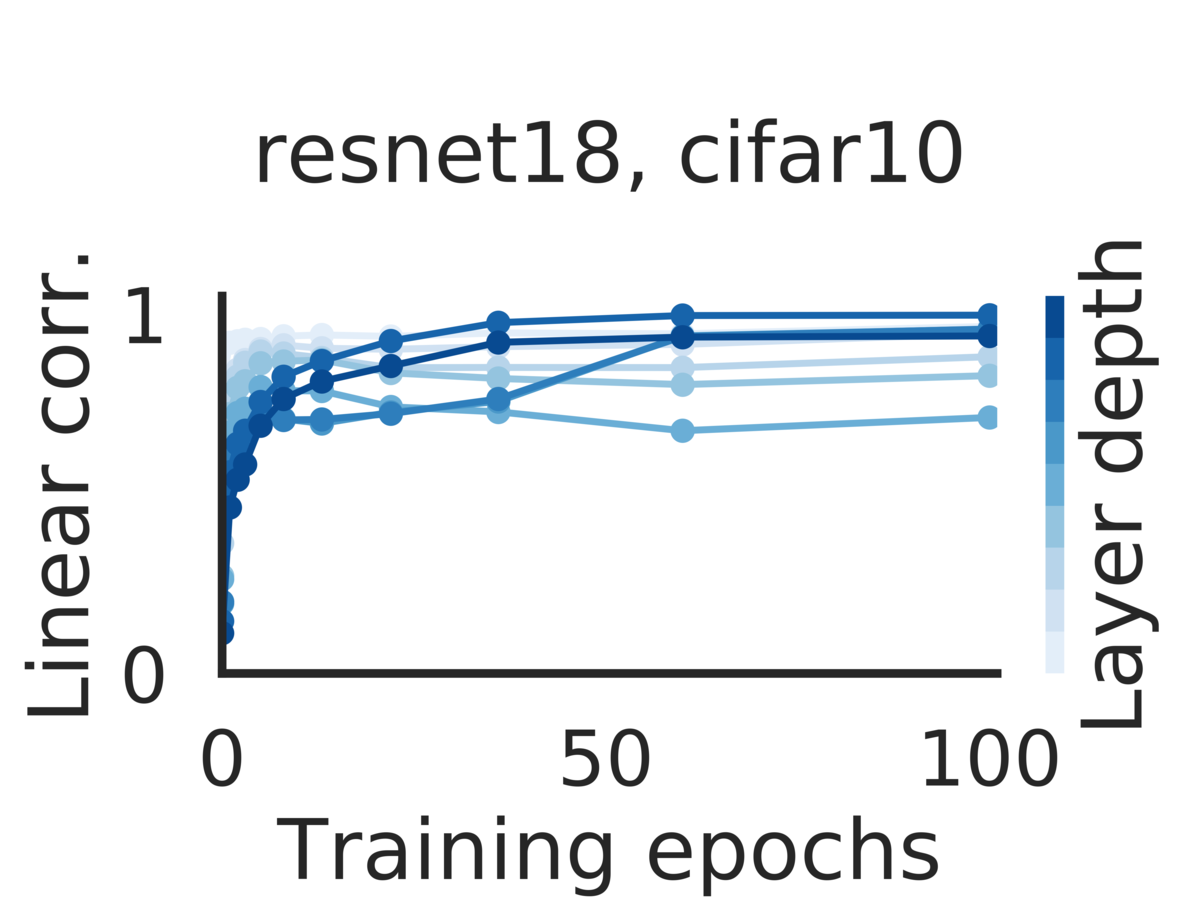}
  \end{minipage}
  \begin{minipage}[b]{0.24\linewidth}
    \centering
    \includegraphics[width=\linewidth]{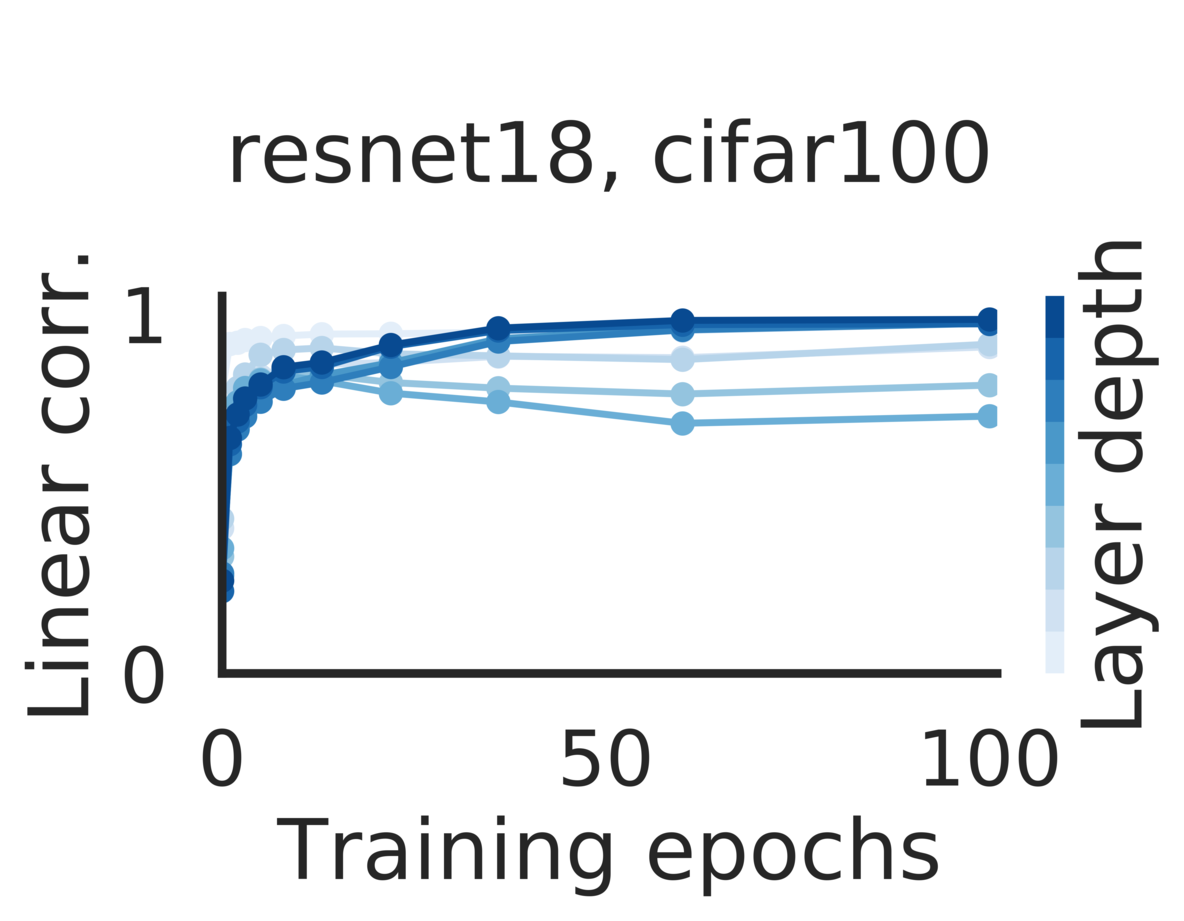}
  \end{minipage}
  
  \begin{minipage}[b]{0.22\linewidth}
    \centering
    \includegraphics[width=\linewidth]{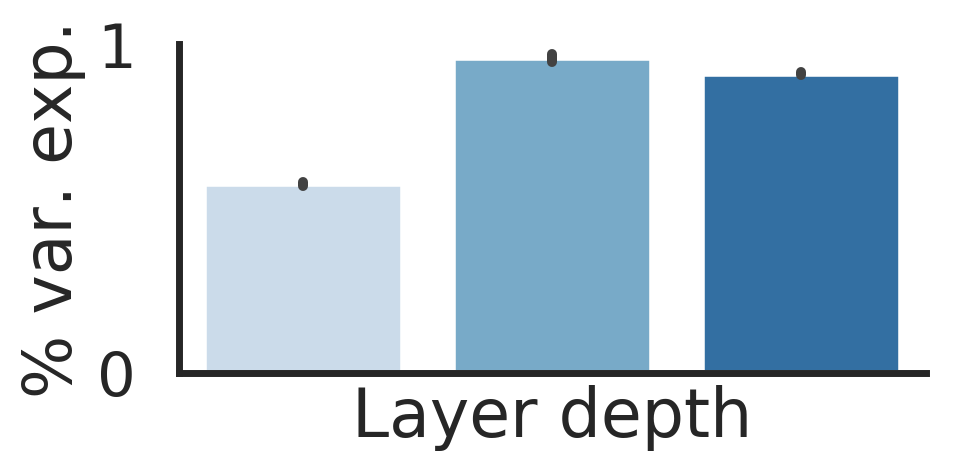}
  \end{minipage}
  \hspace{5pt}
  \begin{minipage}[b]{0.22\linewidth}
    \centering
    \includegraphics[width=\linewidth]{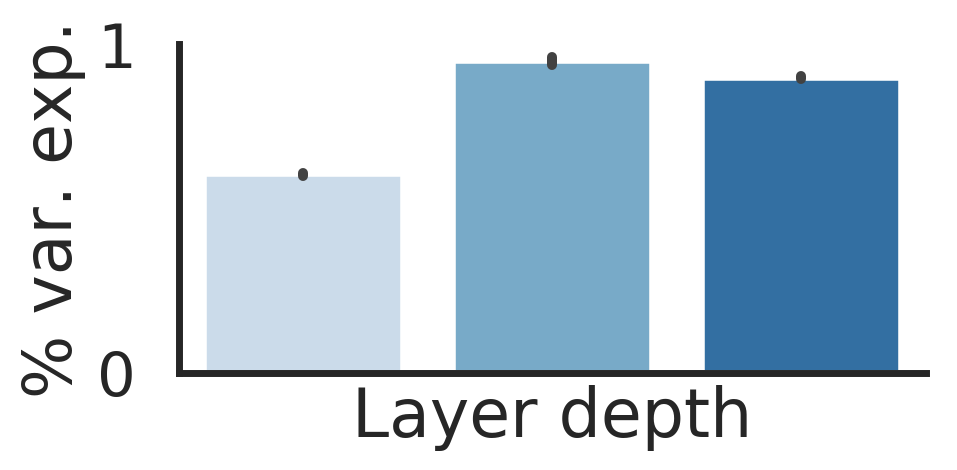}
  \end{minipage} 
  \hspace{5pt}
  \begin{minipage}[b]{0.22\linewidth}
    \centering
    \includegraphics[width=\linewidth]{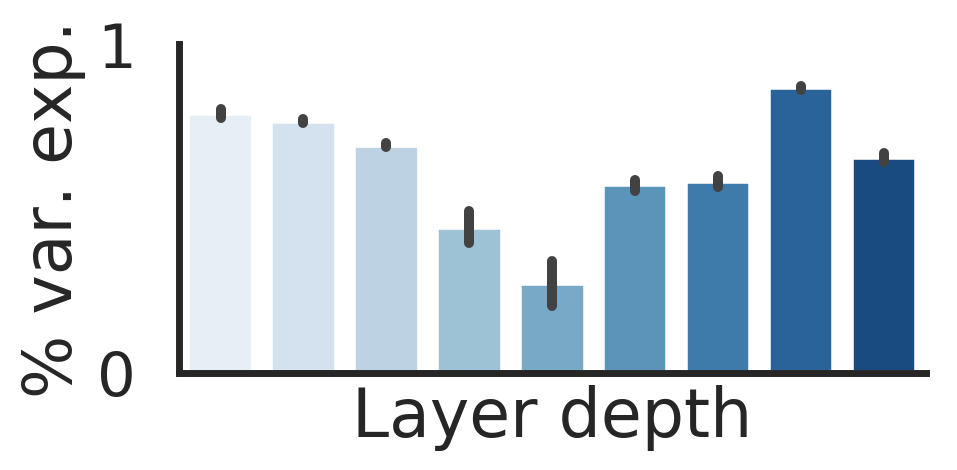}
  \end{minipage}
  \hspace{5pt}
  \begin{minipage}[b]{0.22\linewidth}
    \centering
    \includegraphics[width=\linewidth]{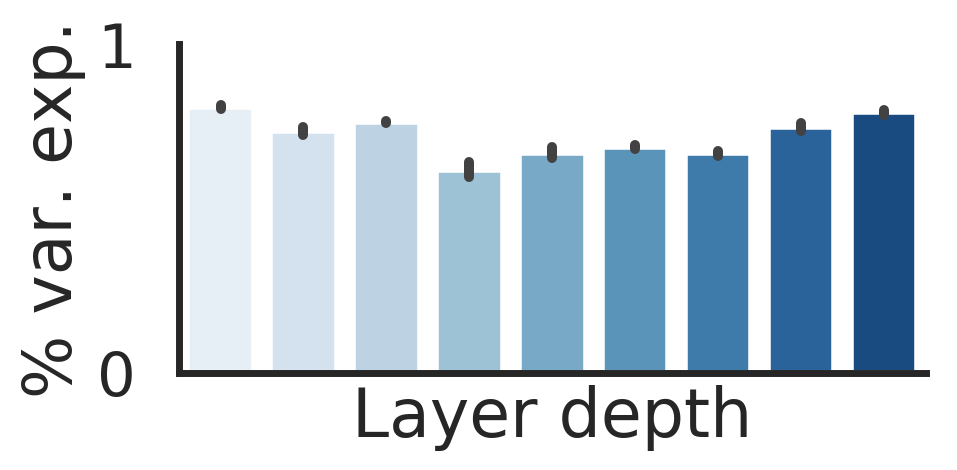}
  \end{minipage} 
  
  \begin{minipage}[b]{0.22\linewidth}
    \centering
    \includegraphics[width=\linewidth]{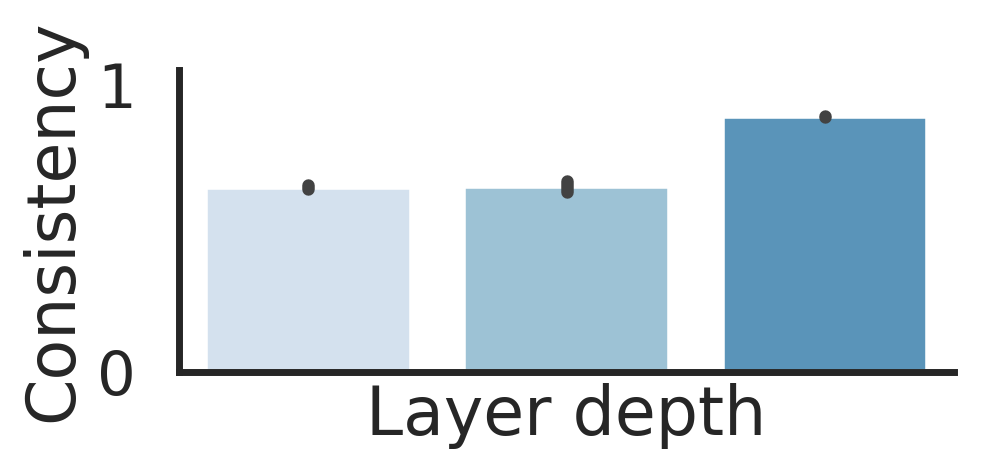}
  \end{minipage}
  \hspace{5pt}
  \begin{minipage}[b]{0.22\linewidth}
    \centering
    \includegraphics[width=\linewidth]{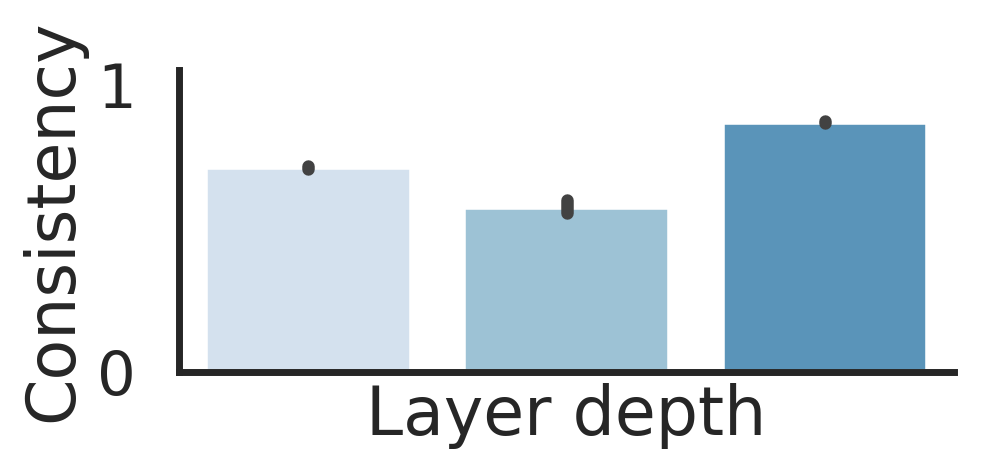}
  \end{minipage} 
  \hspace{5pt}
  \begin{minipage}[b]{0.22\linewidth}
    \centering
    \includegraphics[width=\linewidth]{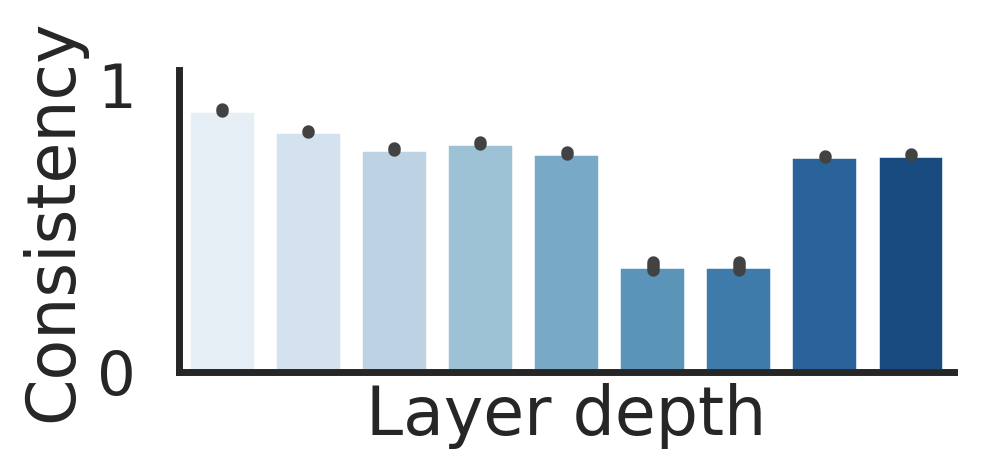}
  \end{minipage}
  \hspace{5pt}
  \begin{minipage}[b]{0.22\linewidth}
    \centering
    \includegraphics[width=\linewidth]{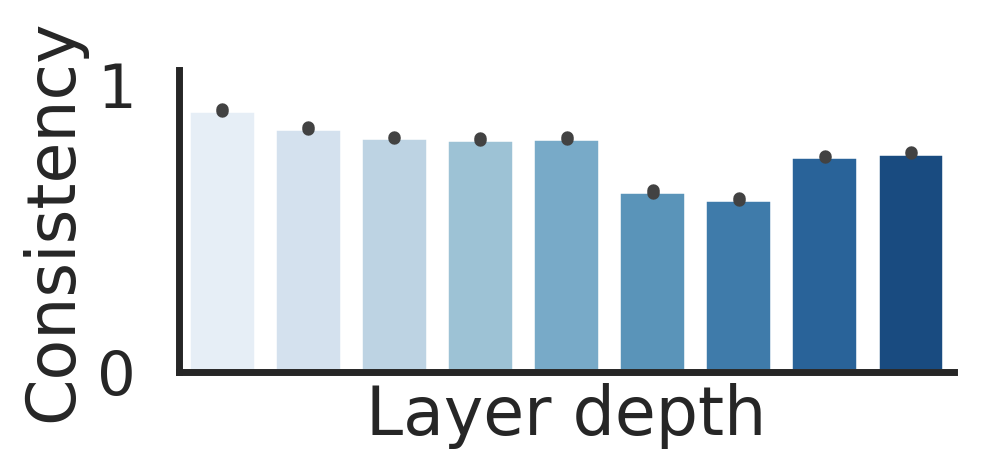}
  \end{minipage}   
\caption{
Evaluate SRM alignment for standard convolutional neural networks and residual networks trained on CIFAR10/CIFAR100.
\textbf{Top row}: The correlation between final averaged within-network RSM and inter-network RSM (in the shared space) during training. Later layers are colored with darker blue; 
\textbf{Middle row}: Proportion variance explained by SRM (on the test set) at the end of the training. 
\textbf{Bottom row}: The average correlation between within-network RSMs. 
Error bars indicate 95\% bootstrapped confidence intervals. 
}
\label{fig-exp}

\end{figure}

\textbf{Experiment: Use SRM to identify the shared representation across neural networks.}
Though the connections between different neural networks are certainly not exactly orthogonal, we will empirically show that orthogonal transformations can provide very accurate alignment across networks. We trained 10 ConvNets and ResNet18 \cite{He2016-eb} on CIFAR10 and CIFAR100 (40 networks in total). The ConvNets have two convolutional layers and one densely connected layer before the output layer. We chose every other layer (9 out of 18) from ResNet18. All networks are trained until convergence. We then recorded their hidden activity patterns on some unseen images, and divided them into a SRM-alignment set and a test set. We used the activity patterns from SRM-alignment set to fit SRM, and then applied to estimated SRM to transform the test set patterns. All later analyses were conducted on the test set. 

Across all experiments, the correlation between average inter-network RSM and average within-network RSM increases through training (Fig.\ref{fig-exp}, top row). This suggests that different neural networks gradually converge to different orthogonal transformations of the same shared representation. 
On the other hand, in the native space, inter-network RSMs do not show any meaningful structure because they are misaligned. 
Figure \ref{fig-demo}B shows examples for within-network RSM, inter-network RSM in the shared space and inter-network RSM in the native space, calculated from ten ConvNets trained on CIFAR10.

The middle row of Fig. \ref{fig-exp} shows the fraction of variance explained by SRM for each layer. For ConvNets, SRM variance explained increases from early layers to deep layers. The proportion variance explained for the last layer is very high (91\% for CIFAR10 and 89\% for CIFAR100), which shows that the activity patterns at the last layer of different ConvNets are roughly learning different orthogonal transformations of an underlying shared representation. 
It is also clear that orthogonal transformation does not fully account for the differences across networks -- variance explained is lower for early ConvNet layers and 
ResNet18. This suggests that early layers and high capacity models have more degrees of freedom to ``choose'' qualitatively different representations. This hypothesis is supported by the consistency across networks (Fig. \ref{fig-exp}, bottom row), measured by average correlation between within-network RSMs (averaged across all pairwise comparisons).

\section{Discussion}

In this work, we found that orthogonal transformation is a good explanation of the differences across different networks independently trained on the same data, which means the geometry of different neural network representations is the same. 
In our experiments, different instances of trained ConvNets were well aligned with orthogonal transformations. 
Orthogonal transformations also explained a large amount of variance for different instances of ResNet18, though the alignment is not as good as ConvNets, suggesting high capacity models might learn qualitatively different representation. 


Why might different networks learn different orthogonal transformations of the same underlying
representation? We think this is a direct consequence of the common input (stimuli) and objective function.
Concretely, for classification tasks, the last hidden layer needs to embed the input examples into some hidden representational space that is decodable by the classifier parameterized by the output layer, and all orthogonal transformations of that representational space are equivalent for the purpose of classification. Interestingly, orthogonal transformations were able to align the responses within each layer along the networks’ hierarchy, suggesting that the nonlinear transformations are similar across networks. 
More generally, we think the shared representational structure across individuals came from the shared experience and shared goals (e.g. object recognition). 
Interestingly, orthogonal transformations revealed shared neural responses across many areas in the visual and auditory cortices as well as other high order cortical areas of biological neural networks as they process real life information \cite{Chen2017-wi, Chen2015-mi, Guntupalli2016-so, Haxby2011-uf}. 
Studying this shared representation might shed light on the intrinsic structure of the shared experience, as well as general properties of neural network representation abstracted away from any specific instances of trained networks.

Here, we studied the relation among neural networks of the same architecture.  However, neural networks with different architectures can also learn to represent the same function with similar representational geometry. For example, the RSMs of high-level visual regions in humans, monkeys and convolutional neural networks are highly similar \cite{Khaligh-Razavi2014-ch}. 
An important future direction will be to understand what's invariant across learned neural representations with different architectures, for both artificial and biological networks.


\subsubsection*{Code \& Demo}

Code repo: \url{https://github.com/qihongl/nnsrm-neurips18}

The shared response model is implemented in BrainIAK: \url{http://brainiak.org/}


\subsubsection*{Acknowledgments}
We thank Andrew M. Saxe and Jonathan D. Cohen for the constructive feedback.
Research reported in this publication was supported by
the NIH Common Fund, the Eunice Kennedy Shriver National Institute of Child Health and Human Development (NICHD) and the NIH Commons Fund, through the Office of Strategic Coordination/Office of the Director (OD), of the National Institutes of Health under award number DP1HD091948 to U.H.
and 
a Multi-University Research Initiative grant to K.A.N. and U.H. (ONR/DoD N00014-17-1-2961). Any opinions, findings, and conclusions or recommendations expressed in this material are those of the author(s) and do not necessarily reflect the views of the Office of Naval Research or the U.S. Department of Defense.

\newpage
\bibliographystyle{plain}
\bibliography{nips_2018}

\end{document}